\documentclass[journal]{IEEEtran}
\usepackage{booktabs}
\usepackage{adjustbox} 
\usepackage{multirow}
\usepackage{amsmath}
\usepackage{booktabs}
\usepackage{xcolor}
\usepackage{amssymb}
\usepackage{graphicx} 
\usepackage{marvosym}
\usepackage{hyperref}

\ifCLASSINFOpdf
\else
\fi
\begin{document}
%
\title{LR-IAD:Mask-Free Industrial Anomaly Detection with Logical Reasoning}
%
%
%


\author{
\IEEEauthorblockN{
Peijian Zeng$^{1}$, 
Feiyan Pang$^{1}$,
Zhanbo Wang$^{1}$,
Aimin Yang$^{2,*}$%
}

\IEEEauthorblockA{
\textit{School of Computer Science and Technology, Guangdong University of Technology, Guangzhou, China}\\
\textit{School of Computer Science and Intelligence Education, Lingnan Normal University, Zhangjian, China}\\
amyang@gdut.edu.cn
}

\thanks{\textsuperscript{*} A. Yang is co-corresponding author.}
}

\maketitle
\begin{abstract}
Industrial Anomaly Detection (IAD) is critical for ensuring product quality by identifying defects. Traditional methods such as feature embedding and reconstruction-based approaches require large datasets and struggle with scalability. Existing vision-language models (VLMs) and Multimodal Large Language Models (MLLMs) address some limitations but rely on mask annotations, leading to high implementation costs and false positives. Additionally, industrial datasets like MVTec-AD and VisA suffer from severe class imbalance, with defect samples constituting only 23.8\% and 11.1\% of total data respectively. To address these challenges, we propose a reward function that dynamically prioritizes rare defect patterns during training to handle class imbalance. We also introduce a mask-free reasoning framework using Chain of Thought (CoT) and Group Relative Policy Optimization (GRPO) mechanisms, enabling anomaly detection directly from raw images without annotated masks. This approach generates interpretable step-by-step explanations for defect localization. Our method achieves state-of-the-art performance, outperforming prior approaches by 36\% in accuracy on MVTec-AD and 16\% on VisA. By eliminating mask dependency and reducing costs while providing explainable outputs, this work advances industrial anomaly detection and supports scalable quality control in manufacturing. Code to reproduce the experiment is available at \href{https://github.com/LilaKen/LR-IAD}{https://github.com/LilaKen/LR-IAD}.
\end{abstract}

\begin{IEEEkeywords}
Industrial Anomaly Detection, Mask-free Reasoning, Multimodal Model
\end{IEEEkeywords}

%
\IEEEpeerreviewmaketitle

\section{Introduction}

Industrial Anomaly Detection (IAD) plays a critical role in visual inspection systems by identifying and localizing defects to ensure product quality. Traditional IAD methods have focused on detecting deviations from normal data, categorized into two primary approaches: \textit{feature embedding-based} and \textit{reconstruction-based} methods. Feature embedding techniques model latent representations of normal samples and use distance metrics for anomaly detection, while reconstruction-based approaches compute reconstruction errors between input and regenerated samples to identify anomalies~\cite{Rudolph_2022_WACV,Yi_2020_ACCV,ZAVRTANIK2021107706}. Despite their effectiveness, these methods require large training datasets, limiting scalability in dynamic industrial environments.

Recent advancements in vision-language models (VLMs), such as CLIP~\cite{pmlr-v139-radford21a}, have introduced paradigm shifts in IAD. Models like AnomalyCLIP~\cite{zhou2024anomalyclip}, PromptAD~\cite{Li_2024_WACV}, and AprilGAN~\cite{chen2023april} leverage CLIP's pre-trained semantic understanding for few-shot and zero-shot anomaly detection. However, their reliance on predefined anomaly concepts restricts generalization to novel defect types. To address this, Multimodal Large Language Models (MLLMs) have emerged as a promising direction. Early examples, such as AnomalyGPT~\cite{Gu_Zhu_Zhu_Chen_Tang_Wang_2024}, demonstrate the feasibility of training MLLMs on IAD datasets. The latest work, VMAD~\cite{deng2024vmad}, integrates fine-grained visual perception with multimodal learning, achieving strong zero-shot performance on benchmarks like MVTec-AD~\cite{Bergmann_2019_CVPR} and VisA~\cite{10.1007/978-3-031-20056-4_23} without additional training. VMAD enhances anomaly localization and provides interpretable explanations through mechanisms like Defect-Sensitive Structure Learning and Locality-Enhanced Token Compression.

\begin{figure}[htbp] 
\centering 
\includegraphics[width=0.5\textwidth]{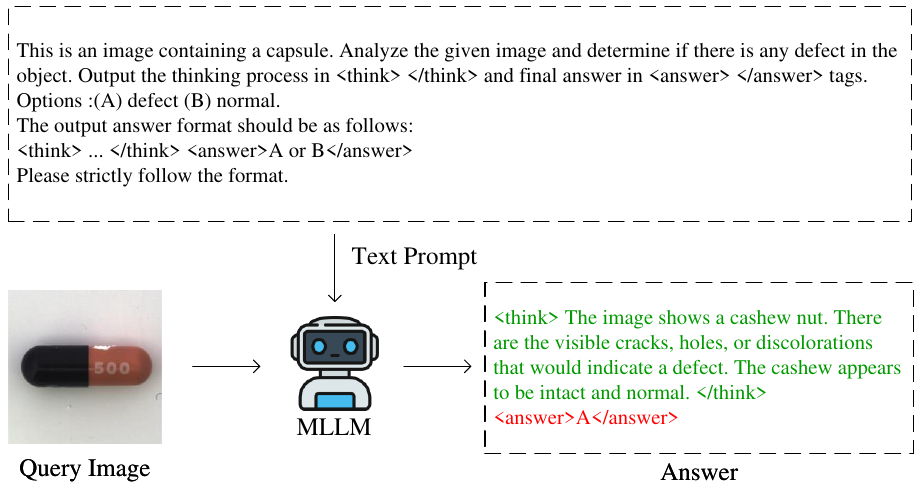} 
\caption{Overview of the task definition.} 
\label{fig:anomaly_example} 
\end{figure}

However, previous approaches to industrial anomaly detection, including AprilGAN~\cite{chen2023april} and AnomalyGPT~\cite{Gu_Zhu_Zhu_Chen_Tang_Wang_2024}, fundamentally rely on explicit mask annotations for training. Real-world methods such as AprilGAN require ground-truth mask annotations for anomalous samples, which are costly and labor-intensive to acquire, severely limiting scalability in large-scale industrial deployments. Meanwhile, synthetic approaches like AnomalyGPT sidestep manual annotation by generating synthetic anomalies from normal samples. However, this strategy introduces its own challenges, the model excessively detects anomalies (over-recall), leading to high false-positive rates. Regardless of their approach—whether relying on real-world annotations or synthetic data—all prior methods share a critical flaw, their dependency on mask-dependent training causes a tendency to misclassify normal samples as defective. In industrial settings, such misclassification is highly detrimental—it triggers unnecessary manual re-inspections, inflates operational costs, and erodes trust in the system's reliability due to frequent false alarms. These limitations collectively underscore the urgent need for mask-free frameworks that maintain robust performance without explicit reliance on annotated masks.

In addition, the success of these methods is challenged by the inherent data imbalance in real-world industrial datasets. As shown in Table~\ref{tab:dataset_imbalance}, popular benchmarks like MVTec-AD and VisA exhibit significant class imbalance—defect samples constitute only 23.8\% and 11.1\% of total samples, respectively. This imbalance exacerbates the difficulty of training models to detect rare defects without overfitting to the dominant normal class, necessitating novel solutions.

Recent advancements in reasoning models, such as Deepseek-R1~\cite{guo2025deepseek}, GRPO~\cite{shao2024deepseekmath}, and (CoT)~\cite{NEURIPS2022_9d560961}, have unlocked new possibilities for IAD. Their core strength lies in reasoning mechanisms that enable two transformative advancements, first, they provide interpretable detection by logically linking visual patterns to defect descriptions, allowing users to understand why a sample is classified as anomalous. Second, these models eliminate the need for costly mask annotations by directly inferring defects from raw visual inputs, reducing reliance on labor-intensive manual labeling. This dual capability not only enhances trust in model decisions through explainable outputs but also lowers implementation costs, making anomaly detection scalable for diverse industrial scenarios. By combining reasoning-driven interpretability with mask-free inference, these models address longstanding limitations of traditional methods, positioning them as a promising foundation for autonomous quality control in modern manufacturing.

\begin{table}[h]
\centering
\caption{Imbalanced Dataset Statistics for Industrial Anomaly Detection}
\label{tab:dataset_imbalance}
\begin{tabular}{lccc}
\toprule
\textbf{Dataset} & \textbf{Defect Samples} & \textbf{Normal Samples} & \textbf{Defect Ratio (\%)} \\
\midrule
MVTec-AD & 1,258 & 4,096 & 23.8\% \\
VisA & 1,200 & 9,621 & 11.1\% \\
\bottomrule
\end{tabular}
\end{table}

Motivated by these developments and the identified gap, we summarize our contributions as follows:

\begin{enumerate}
\item \textbf{A novel reward function for imbalanced data}: We propose a reward mechanism that dynamically prioritizes rare defect patterns during training, addressing class imbalance in industrial datasets without overfitting to majority classes.
\item \textbf{Mask-free reasoning via Chain of Thought}: Our framework eliminates dependency on annotated masks by leveraging Chain of Thought reasoning to infer anomalies directly from raw images. This approach generates interpretable step-by-step explanations (e.g., visualizing defect localization and logical inference paths).
\item \textbf{State-of-the-art performance in zero-shot settings}: We outperform the best prior method by \textbf{+36 points} in accuracy on MVTec-AD and \textbf{+16 points} on VisA.
\end{enumerate}

These contributions address some challenges in industrial anomaly detection, including data imbalance and mask dependency.

\section{Related Work}

\subsection{Traditional Anomaly Detection Methods}

Traditional industrial anomaly detection methods have historically relied on classical image processing and statistical techniques to identify manufacturing defects~\cite{liu2024deep, 10443076}. The advent of few-shot learning marked a shift toward addressing data scarcity by enabling models to generalize from limited labeled samples~\cite{KAMOONA2024107706, Wu_2021_ICCV}. Meta-learning approaches, such as~\cite{10.1007/978-3-031-20053-3_18, Wu_2021_ICCV}, depend on extensive meta-training for adaptation, whereas methods like PatchCore~\cite{Roth_2022_CVPR}, SPADE~\cite{cohen2020sub}, and PaDiM~\cite{10.1007/978-3-030-68799-1_35} operate with minimal support sets but lack specific optimization for few-shot anomaly detection. Despite advancements, these techniques remain constrained by inefficiency and limited adaptability in complex industrial environments.

To reduce reliance on labeled data, zero-shot anomaly detection has emerged as a promising direction, leveraging large-scale pre-trained models for broader generalization. For instance, MAEDAY~\cite{SCHWARTZ2024103958} employs a masked autoencoder~\cite{He_2022_CVPR} for reconstruction-based anomaly localization. Meanwhile, CLIP-based methods such as AprilGAN~\cite{chen2023april}, WinCLIP~\cite{Jeong_2023_CVPR}, AdaCLIP~\cite{10.1007/978-3-031-72761-0_4}, and AnomalyCLIP~\cite{zhou2024anomalyclip} integrate VLMs with feature matching strategies. MuSc~\cite{li2024musc} introduces a novel zero-shot method that utilizes unlabeled test images through a Mutual Scoring Mechanism, effectively distinguishing anomalies by leveraging implicit normal and abnormal cues. However, most models still lack contextual reasoning capabilities, often relying on mismatches in pre-trained patterns rather than causal analysis.

\subsection{MLLMs-Based Anomaly Detection and Generalization}

The integration of Multimodal Large Language Models (MLLMs) into IAD has emerged as a dynamic and rapidly evolving area of research in recent years~\cite{yang2025survey, jiang2024mmad}. Among the early contributions, AnomalyGPT~\cite{Gu_Zhu_Zhu_Chen_Tang_Wang_2024} advanced the field by adapting MLLMs to interpret feature maps from expert models, achieving strong zero-shot anomaly detection performance in industrial settings. Myriad~\cite{li2023myriad} established a foundational framework by combining large language models (LLMs) with vision expert models. This pioneering approach introduced a classical structure that significantly influenced subsequent studies focused on integrating visual and linguistic processing. However, Myriad's reliance on carefully curated vision expert models introduces complexity, potentially limiting scalability in diverse industrial applications. Meanwhile, Echo~\cite{chen2025can} introduced a collaborative framework where specialized MLLMs work together, enhancing detection through system-level synergy, though it avoids full fine-tuning for IAD tasks.In contrast, LogicAD~\cite{Jin_Feng_Mou_Lakemeyer_Decker_Simons_Stegmaier_2025} and LogiCode~\cite{10710633} approached anomaly detection through logical reasoning, offering a unique perspective that excels when anomalies are defined in rational terms. 

Other notable works include~\cite{zhang2025eiad}, which employs Supervised Fine-Tuning but struggles to achieve significant growth in MMAD benchmarks~\cite{jiang2024mmad}. Anomaly-R1~\cite{chao2025anomalyr1} standed out by leveraging a compact MLLM enhanced with ROAM-guided GRPO, enabling end-to-end anomaly detection even in scenarios with scarce defect data. Similarly, LAD-Reasoner~\cite{li2025lad} proposed a two-stage framework for logical anomaly detection, achieving excellent interpretability and performance while maintaining a small model size. These advancements collectively highlight the ongoing innovation in MLLM-driven IAD, addressing challenges such as scalability, adaptability, and interpretability in real-world industrial contexts.

\section{Method}
We introduce Logical Reasoning for Industrial Anomaly Detection (LR-IAD) , a novel framework designed to enhance anomaly detection performance through a mask-free, multimodal approach. Our method leverages the integration of visual and textual inputs, utilizing the Qwen2-VL 7B~\cite{wang2024qwen2vlenhancingvisionlanguagemodels} model as the baseline architecture. To further refine the model's capabilities, we incorporate GRPO~\cite{shao2024deepseekmath}, which optimizes the reasoning process for detecting anomalies.

The learning process in LR-IAD is guided by two reward functions: format reward , which ensures structured and interpretable outputs, and focal reward , inspired by the focal loss mechanism~\cite{Lin_2017_ICCV}. The focal reward is designed to emphasize the identification of critical anomaly regions by mitigating the dominance of easy-to-detect normal samples and focusing on hard-to-detect anomalies. By combining these components, LR-IAD achieves robust and scalable anomaly detection without relying on annotated masks, addressing key limitations in traditional industrial anomaly detection methods. Our framework offers a scalable, interpretable solution (as shown in Figure~\ref{fig:anomaly_example}), enabling high-performance quality control in real-world manufacturing scenarios.

\begin{figure*}[htbp] 
    \centering 
    \includegraphics[width=1\textwidth]{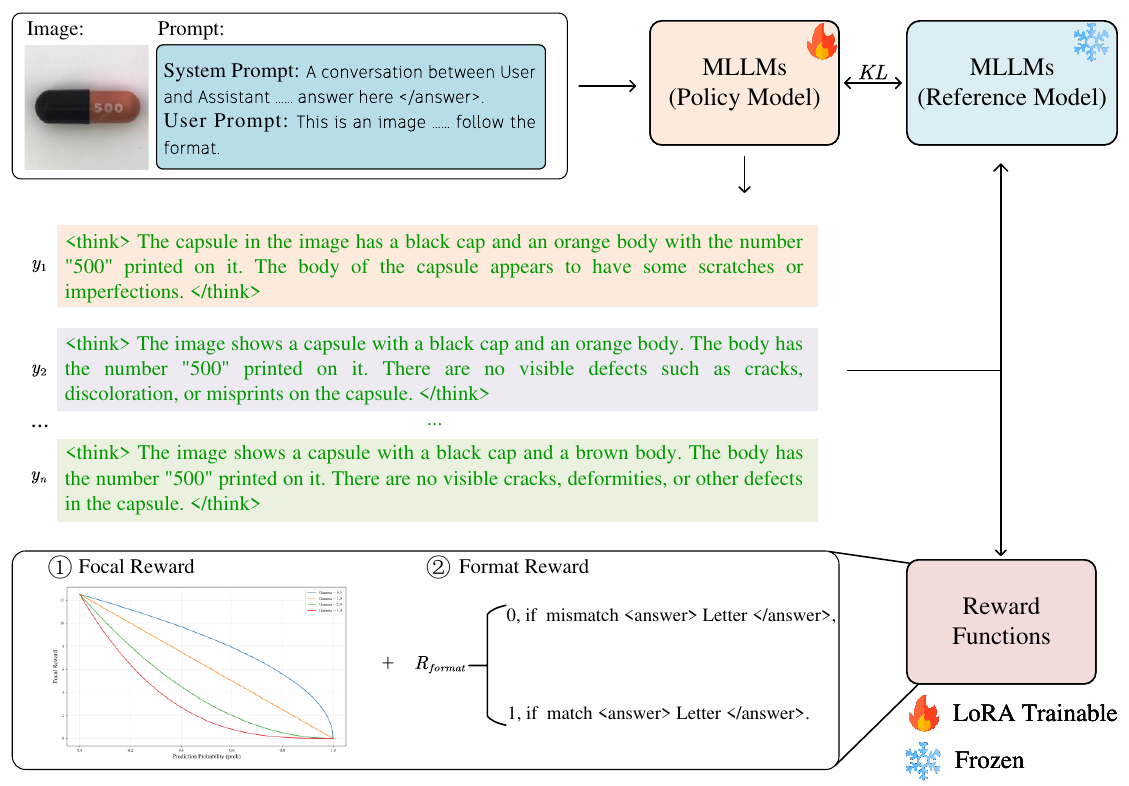} 
    \caption{Overview of the LR-IAD framework, illustrating its components and reasoning process for anomaly detection.} 
    \label{fig:anomaly_example} 
\end{figure*}

\subsection{Problem Definition}

The problem addressed in this work is to detect anomalies in industrial images without relying on annotated masks. Given a image and a textual prompt describing the object of interest, the MLLMs must analyze the image and determine whether the object contains any defects. Mathematically, the input can be represented as

\begin{equation}
I = \{x_{i}, t_{i}\}_{i=1}^{N},
\end{equation}
where \(x_{i}\) represents the image data for the \(i\)-th sample, \(t_{i}\) represents the corresponding textual prompt, and \(N\) is the total number of samples in the dataset. The textual prompt is formulated as

\begin{equation}
t_{i} = f(o_{i}),
\end{equation}
where \(f(\cdot)\) is a function that generates a structured query based on the object name \(o_{i}\). For example, given an object name like "pill," the textual prompt might include descriptions such as "identify anomalies in the pill." Images are preprocessed to ensure consistent resolution and normalization, while textual prompts are tokenized into a unified format suitable for input into the MLLMs. By leveraging raw visual and textual inputs without requiring annotated masks, our approach significantly reduces the dependency on labor-intensive labeling processes.

\subsection{Reward Functions}

To optimize the model's performance, we introduce two reward functions that guide the learning process: format reward and focal reward.

\subsubsection{Format Reward}

The format reward ensures that the model's output adheres to a predefined structure, penalizing deviations from the expected format. It maintains consistency in anomaly detection outputs. The reward is formulated as

\begin{equation}
R_{format} = 
\begin{cases} 
1, & C_{1}, \\
0, & C_{2}.
\end{cases}
\end{equation}
Here, \(C_{1}\) represents the condition indicating that the output satisfies the specified format, and \(C_{2}\) represents the condition indicating that the output deviates from the expected structure. By enforcing structured outputs, the format reward improves usability in industrial applications.

\subsubsection{Focal Reward}

The focal reward is inspired by the focal loss mechanism~\cite{Lin_2017_ICCV}. It emphasizes hard-to-classify samples, ensuring that the MLLMs focuses on challenging cases rather than easy-to-detect normal samples. The focal reward is computed as

\begin{equation}
R_{focal} = \alpha \cdot (1 - p)^{\gamma} \cdot f,
\end{equation}
where \(p\) is the predicted probability of the correct answer, \(\alpha\) is the scaling factor controlling the weight of the reward, \(\gamma\) is the focusing parameter to adjust the emphasis on hard samples (higher values increase focus on low-confidence predictions), and \(f\) is the scaling factor to amplify the reward for critical anomaly regions. This mechanism ensures that low-confidence misclassified samples receive higher penalties, guiding the model to prioritize difficult cases and improve overall robustness.

\subsection{GRPO Methodology}

Group Relative Policy Optimization (GRPO) is a key component of our framework enabling the model to learn optimal policies for anomaly detection through reinforcement learning. GRPO combines policy evaluation advantage estimation and policy updates to iteratively refine the model's decision-making process. Below we provide a detailed explanation of each step in the GRPO methodology.

Policy evaluation involves generating predictions for each input sample \(I_{i} = \{x_{i}, t_{i}\}\) which are evaluated using the reward functions described earlier. The reward for the \(t\)-th sample is calculated as

\begin{equation}
r_{t} = R(a_{t}, s_{t}),
\end{equation}
where \(a_{t}\) represents the action taken by the model such as anomaly classification or localization \(s_{t}\) represents the state of the environment including the image and textual prompt and \(R(\cdot)\) represents the reward function combining format and focal rewards. The reward signal provides critical feedback on the quality of the model's predictions guiding the optimization process. To ensure stable learning GRPO incorporates a KL divergence term that penalizes large deviations from the current policy during updates. This helps maintain consistency in the model's behavior while allowing gradual improvements.

Advantage estimation measures how much better the current action is compared to the average performance. In GRPO, the advantage function is computed through a normalization process that enhances stability and efficiency during training.

The procedure begins by aggregating rewards from multiple reward functions into a single scalar value for each sample:
\begin{equation}
r = \sum_{i} r_i,
\end{equation}
where \(r_i\) represents the reward from the \(i\)-th reward function.

Next, the rewards are divided into groups, and the mean and standard deviation of each group are calculated:
\begin{equation}
\mu_r = \frac{1}{N} \sum_{i=1}^{N} r_i, \quad \sigma_r = \sqrt{\frac{1}{N} \sum_{i=1}^{N} (r_i - \mu_r)^2},
\end{equation}
where \(N\) is the number of samples in each group.

The advantage for each sample is then computed by normalizing the rewards within their respective groups:
\begin{equation}
A_t = \frac{r_t - \mu_r}{\sigma_r + \epsilon},
\end{equation}
where \(\epsilon\) is a small constant added to prevent division by zero. This normalization ensures that the advantage values have zero mean and unit variance, reducing numerical instability and improving training robustness.

To further stabilize the training process, GRPO incorporates the Kullback-Leibler (KL) divergence between the old and new policies into the optimization. This regularization ensures that policy updates remain within a trust region, preventing abrupt changes that could destabilize training. By combining normalized advantages with KL divergence regularization, GRPO achieves a balance between optimizing for higher rewards and maintaining policy stability, leading to more reliable and efficient learning.

Policy updates adjust the model parameters \(\theta\) using gradient ascent to maximize the expected reward while minimizing the KL divergence between the old and new policies. The update rule is given by

\begin{equation}
\theta \leftarrow \theta + \eta \nabla_{\theta} \mathbb{E} \left[ \sum_{t} A_{t} \log \pi_{\theta}(a_{t} | s_{t}) - \beta \cdot D_{\text{KL}}(\pi_{\text{old}} || \pi_{\theta}) \right],
\end{equation}
where \(\eta\) is the learning rate controlling the step size of parameter updates, \(\pi_{\theta}(a_{t} | s_{t})\) is the probability of taking action \(a_{t}\) in state \(s_{t}\) under the current policy, \(D_{\text{KL}}(\pi_{\text{old}} || \pi_{\theta})\) is the KL divergence between the old policy,\(\pi_{\text{old}}\) and the updated policy, \(\pi_{\theta}\) and \(\beta\) is the regularization coefficient controlling the strength of the KL divergence penalty. This update rule ensures that the model learns to take actions that maximize the cumulative reward while maintaining stability through KL divergence regularization. By balancing exploration and exploitation GRPO enables the model to achieve robust generalization across diverse industrial anomaly detection tasks.

\subsection{Baseline Model and Prompt Template}

We use Qwen2-VL~\cite{wang2024qwen2vlenhancingvisionlanguagemodels} as the baseline model, which is fine-tuned using GRPO. The model is deployed across multiple GPUs using multiprocessing to handle large-scale datasets efficiently. To guide the model's reasoning process, we employ the following prompt template:

\begin{quote}
\footnotesize
"This is an image containing a \texttt{object\_name}. Analyze the given image and determine if there is any defect in the object.\newline
Output the thinking process in \texttt{<think>} \texttt{</think>} and final answer in \texttt{<answer>} \texttt{</answer>} tags.\newline
Options: (A) defect (B) normal.\newline
The output answer format should be as follows:\newline
\texttt{<think>} ... \texttt{</think>} \texttt{<answer>A or B</answer>}\newline
Please strictly follow the format."
\end{quote}

This template ensures that the model generates interpretable outputs by explicitly separating the reasoning process (enclosed in ``\texttt{<think>}'' tags) from the final decision (enclosed in ``\texttt{<answer>}'' tags). By enforcing strict formatting rules, the model produces structured and reliable results, enhancing its usability in real-world industrial applications.

\subsection{Inference Pipeline}

During inference, the model processes each input sample \(I_{i} = \{x_{i}, t_{i}\}\) and generates a response. The response is parsed to extract the predicted label, which is compared against the ground truth label. The prediction process can be summarized as

\begin{equation}
y_{i} = g(x_{i}, t_{i}; \theta),
\end{equation}
where \(g(\cdot)\) is the prediction function parameterized by \(\theta\), and \(y_{i}\) is the predicted label for the \(i\)-th sample. The inference pipeline is fully mask-free, making it scalable and adaptable to real-world industrial scenarios where annotated masks are unavailable or impractical to obtain.

\section{Experiment}

\subsection{Experiment Setting}

All experiments were conducted on a server with four NVIDIA A40 GPUs, each equipped with 46GB of memory. The training batch size was set to 8, calculated as 1 sample per GPU multiplied by 4 GPUs and 2 gradient accumulation steps. The model was trained for 1 epoch, with checkpoints saved every 100 steps. Low-Rank Adaptation (LoRA)~\cite{hu2022lora} was employed for efficient fine-tuning using parameters such as a rank of 4, an $\alpha$ value of 16, a dropout rate of 0.1, and target modules $Q$ and $V$. Key configurations included a maximum prompt length of 1024 tokens, gradient checkpointing disabled, and the Flash Attention 2~\cite{dao2024flashattention} implementation for attention mechanisms. The maximum input pixel capacity was set to 401,408 (equivalent to approximately $640 \times 640$ resolution), and mixed-precision training was performed using bfloat16. Only the final model weights were saved, with intermediate steps omitted. Distributed training was accelerated using DeepSpeed~\cite{Li_Yao_Wu_Zhang_Holmes_Li_He_2024} ZeRO-3\footnote{Our experiments are facilitated by the Visual-RFT repository: \url{https://github.com/Liuziyu77/Visual-RFT}.}. Our work utilizes PaddlePaddle as the deep learning frameworks\footnote{PaddlePaddle is an open-source deep learning platform with a simple API designed by Baidu.}.

\subsection{Evaluation Metrics}

To evaluate the performance of our model, we utilized two metrics, Accuracy and F1-macro~\cite{SOKOLOVA2009427}. These metrics provide a comprehensive understanding of the model's predictive capabilities.

Accuracy measures the proportion of correctly classified samples out of the total number of samples. It is calculated as follows:

\begin{equation}
    \mathrm{Accuracy} = \frac{\mathrm{TP} + \mathrm{TN}}{\mathrm{TP} + \mathrm{TN} + \mathrm{FP} + \mathrm{FN}}
\end{equation}
where TP represents true positives, TN represents true negatives, FP represents false positives, and FN represents false negatives. Accuracy provides a general sense of the model's correctness but may not fully capture performance in imbalanced datasets.

The F1 score for each class is the harmonic mean of precision and recall, balancing the trade-off between the two. The macro-averaged F1 score (F1-macro) is computed by taking the unweighted average of F1 scores across all classes:

\begin{equation}
    \mathrm{F1\mbox{-}macro} = \frac{1}{N} \sum_{i=1}^{N} \mathrm{F1}_i
\end{equation}
where $N$ is the total number of classes, and $\mathrm{F1}_i$ is the F1 score for the $i$-th class. Macro-averaging ensures that all classes contribute equally to the final score, regardless of their distribution in the dataset. This metric is particularly useful when dealing with multi-class classification tasks where class imbalance exists.

Both metrics are reported to provide a complete picture of the model's effectiveness in terms of overall correctness and class-wise balance.

\subsection{Data Partitioning Strategies}

The data distribution for training and inference across different models is summarized in Table~\ref{tab:model_data_distribution}. This table clarifies how each model leverages the VisA and MVTec-AD datasets, with critical insights into their design limitations.

For AprilGAN, training is restricted to the Test subsets of both datasets due to its reliance on explicit mask annotations. However, this strategy limits training data availability, forcing it to use testing data for model learning—a practice that risks overfitting and undermines evaluation objectivity. During inference, AprilGAN evaluates on the complementary dataset's Test subset (e.g., MVTec-AD for VisA training), but its dependency on mask-dependent training introduces over-recall bias. This leads to excessive anomaly detection, misclassifying normal samples as defective due to overfitting to mask patterns.

AnomalyGPT adopts a different approach by training on the Train subsets of both datasets. While this avoids using Test data for training, its requirement for mask annotations during training (even on Train subsets) forces it to generate synthetic anomalies using the NSA method~\cite{10.1007/978-3-031-19821-2_27}. The NSA method builds upon the Cut-paste~\cite{Li_2021_CVPR} technique by incorporating Poisson image editing~\cite{10.1145/3596711.3596772} to alleviate discontinuities caused by pasting image segments. Cut-paste randomly crops a region from an image and pastes it onto another location, creating artificial anomalies. Although Poisson editing improves realism by solving Poisson partial differential equations to blend regions seamlessly, synthetic generation still introduces noise. This noise causes high false-positive rates during inference, as the model becomes overly sensitive to minor irregularities. As shown in the table, AnomalyGPT evaluates on Test subsets but struggles with precision, particularly in distinguishing subtle normal variations from anomalies.

LR-IAD eliminates mask dependency entirely. By training on the entire dataset (All) for both VisA and MVTec-AD, we leverage all available data without relying on annotated masks. This mask-free strategy avoids over-recall bias and enables robust generalization, as our framework learns to infer anomalies directly from raw visual features.

The Table~\ref{tab:model_data_distribution} highlights the inherent trade-offs: mask-dependent methods (AprilGAN, AnomalyGPT) are constrained by limited data usage and over-sensitivity to annotations, whereas LR-IAD maximizes dataset utilization while mitigating false positives through mask-free reasoning. To ensure consistency across experiments, we set the 0-shot test set to include all samples from the dataset, thereby guaranteeing a fair evaluation of the practicality of each method.

\begin{table}[htbp]
    \centering
    \caption{Training and Inference Data Partitioning Strategies for VisA and MVTec-AD}
    \label{tab:model_data_distribution}
    \begin{tabular}{lcc}
        \toprule
        \textbf{Model} & \textbf{Training Data} & \textbf{Inference Data} \\
        \midrule
        AprilGAN       & VisA (Test)           & MVTec-AD (Test)         \\
                       & MVTec-AD (Test)       & VisA (Test)             \\
        AnomalyGPT     & VisA (Train)          & MVTec-AD (Test)         \\
                       & MVTec-AD (Train)      & VisA (Test)             \\
        LR-IAD           & VisA (All)            & MVTec-AD (All)          \\
                       & MVTec-AD (All)        & VisA (All)              \\
        \bottomrule
    \end{tabular}
\end{table}

\subsection{Main Result}

\begin{table*}[htbp]
    \centering 
    \caption{Performance Comparison of Methods on MVTec-AD and VisA Datasets (0-shot), Bold values indicate the best performance.}
    \label{tab:performance_comparison}
    \resizebox{\textwidth}{!}{ 
    \begin{tabular}{cccccc}
        \toprule
        Setup                   & Method     & Accuracy (MVTec-AD) & F1-marco(MVTec-AD) & Accuracy (VisA) & F1-marco(VisA) \\
        \midrule
        \multirow{3}{*}{0-shot}  
                                & WinCLIP   & 24.47               & 20.43              & 11.09           & 9.98          \\
                                & AprilGAN   & 47.88               & 42.78              & 71.97           & 55.52          \\
                                & AnomalyGPT & 33.12               & 28.51              & 14.64           & 12.91          \\
                                & LR-IAD       & \textbf{84.35}      & \textbf{71.54}     & \textbf{87.60}  & \textbf{59.54} \\
        \bottomrule
    \end{tabular}
    }
\end{table*}

Table~\ref{tab:performance_comparison} presents the zero-shot performance of three methods on the MVTec-AD and VisA datasets. Our proposed LR-IAD achieves significant improvements over baseline models, demonstrating the effectiveness of its mask-free reasoning framework.

AprilGAN shows moderate performance (47.88\% accuracy on MVTec-AD and 71.97\% on VisA), constrained by its reliance on mask annotations for training. Its F1 scores (42.78 on MVTec-AD and 55.52 on VisA) indicate challenges in balancing precision and recall, likely due to overfitting to limited training data.

AnomalyGPT, despite its synthetic anomaly generation capability, performs poorly in both datasets. Its accuracy drops to 33.12\% on MVTec-AD and 14.64\% on VisA, with F1 scores below 30\% for MVTec-AD. This stark underperformance suggests that synthetic data generation introduces noise, overwhelming the model's ability to distinguish real anomalies from artifacts created during training. Moreover, models trained on synthetic anomaly masks suffer from the same issue as AprilGAN, where they can identify anomaly classes but frequently misclassify normal samples as anomalous, further degrading their reliability in practical applications.

In contrast, LR-IAD outperforms all baselines by a large margin. On MVTec-AD, it achieves 84.35\% accuracy and 71.54 F1-score, surpassing AprilGAN by 36.47 percentage points in accuracy and 28.76 points in F1. On VisA, LR-IAD's 87.60\% accuracy and 59.54 F1-score represent 72.96\% improvement over AprilGAN's accuracy and 4.02 point gain in F1 . These results validate the benefits of our mask-free strategy, which avoids overfitting to annotated masks while leveraging chain-of-thought reasoning to generalize across datasets.

The performance gap between MVTec-AD and VisA highlights dataset-specific challenges. While AprilGAN performs better on MVTec-AD (industrial objects), its reliance on Test subset training may limit its adaptability to VisA's more visually complex anomalies. AnomalyGPT's failure on VisA (14.64\% accuracy) underscores the risks of synthetic anomaly generation, which amplifies noise and undermines precision.

Our method's consistent superiority across both datasets demonstrates its robustness to data imbalance and mask dependency. The results align with our design goals: eliminating mask annotations reduces over-sensitivity to artificial patterns, while logical reasoning enhances anomaly localization accuracy.

\subsection{Ablation Study}

To evaluate the effectiveness of our proposed method, LR-IAD, we conducted an ablation study comparing its performance with the baseline model Qwen2-VL (base) in zero-shot settings. The results are summarized in Table~\ref{tab:ablation_study}, which reports key metrics such as Accuracy and F1-macro on two benchmark datasets: MVTec-AD and VisA.

The experimental results demonstrate significant improvements achieved by LR-IAD over the baseline model across all metrics and datasets. On the MVTec-AD dataset, LR-IAD achieves an accuracy of 84.35\%, representing a gain of +14.51\% compared to Qwen2-VL (base)'s 69.84\%. Similarly, the F1-macro score increases from 30.88\% to 71.54\%, reflecting a substantial improvement of +40.66\%. These gains indicate that LR-IAD not only improves overall classification accuracy but also enhances the model's ability to detect rare anomalies, as reflected in the higher F1-macro score.

On the VisA dataset, LR-IAD demonstrates even more pronounced improvements. The accuracy rises from 70.80\% to 87.60\% (+16.80\%), and the F1-macro score improves from 20.56\% to 59.54\% (+38.98\%). These results highlight LR-IAD's robustness and adaptability, especially in handling more complex and diverse data distributions. The VisA dataset includes a broader range of defect types and categories, making it a more challenging benchmark. The significant performance gains on VisA underscore LR-IAD's ability to generalize across diverse industrial scenarios without relying on annotated masks.

One of the key factors contributing to LR-IAD's success is its use of reward functions, including format rewards and focal rewards. These mechanisms ensure that the model prioritizes hard-to-classify samples while maintaining structured outputs. Additionally, LR-IAD's integration of logical reasoning through the \texttt{<think>} and \texttt{<answer>} tags enhances interpretability by explicitly separating the reasoning process from the final decision. This design not only improves trust in the system but also facilitates error analysis and further refinement.

In summary, the ablation study confirms the superiority of LR-IAD over the baseline model Qwen2-VL (base) in zero-shot anomaly detection tasks. The significant improvements in accuracy and F1-macro scores across diverse datasets highlight the effectiveness of our approach in addressing some challenges such as class imbalance, generalization, and interpretability.

\begin{table*}[t]
    \centering
    \caption{Zero-shot performance comparison on MVTec-AD and VisA datasets. Bold values show improvements with absolute gains in parentheses.}
    \resizebox{\textwidth}{!}{ 
        \begin{tabular}{cccccc}
        \toprule
        Setup                   & Method         & Accuracy (MVTec-AD) & F1-marco(MVTec-AD) & Accuracy (VisA) & F1-marco(VisA) \\
        \midrule
        \multirow{2}{*}{0-shot} & Qwen2-VL(base) & 69.84                    & 30.88                   & 70.80               & 20.56               \\
                                & LR-IAD         & \textbf{84.35(+14.51)}      & \textbf{71.54(+40.66)}     & \textbf{87.60(+16.80)}  & \textbf{59.54(+38.98)} \\
        \bottomrule
        \end{tabular}
    }
    \label{tab:ablation_study}
\end{table*}

\subsection{Case Study}

\begin{figure}[htbp] 
    \centering 
    \includegraphics[width=0.5\textwidth]{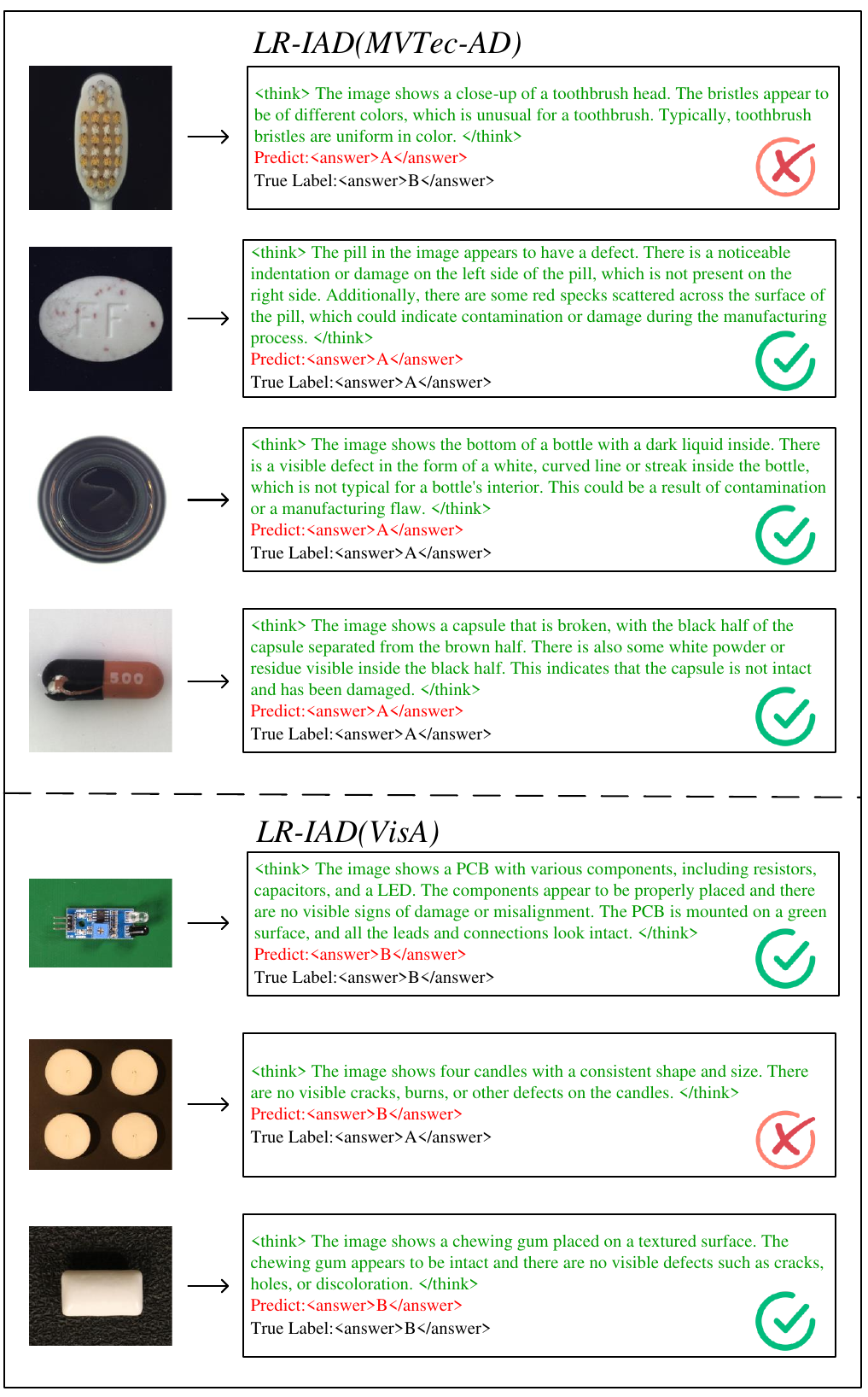} 
    \caption{Visualization examples of anomaly detection results on the MVTec-AD and VisA datasets, showcasing normal and anomalous samples.}
    \label{fig:case_study} 
\end{figure}

To evaluate the practical performance of LR-IAD, we conducted a case study on two benchmark datasets: MVTec-AD and VisA. The results are visualized in Figure~\ref{fig:case_study}, which provides examples of anomaly detection outputs for both normal and anomalous samples. These examples highlight the model's reasoning process and its ability to generate interpretable explanations through structured tags such as \texttt{<think>} and \texttt{<answer>}.

The case study demonstrates LR-IAD's effectiveness in handling diverse industrial scenarios. For instance, in the MVTec-AD dataset, the model correctly identifies anomalies in objects like bottle and capsule by leveraging multimodal reasoning. The \texttt{<think>} tag captures the intermediate reasoning steps, providing insights into how the model analyzes visual features and textual prompts. Similarly, in the VisA dataset, LR-IAD successfully detects complex defects that are often challenging for traditional methods. The structured output format ensures consistency and transparency, enabling users to understand the decision-making process.

One notable example is the detection of a defect in a bottle sample from MVTec-AD. The model generates a detailed reasoning process in the \texttt{<think>} tag, identifying subtle irregularities in the object's surface. The final prediction in the \texttt{<answer>} tag correctly classifies the sample as anomalous (A), matching the ground truth label. Another example from the VisA dataset involves a normal sample where the model accurately predicts the absence of defects (B). These cases illustrate LR-IAD's robustness in balancing precision and recall, even in zero-shot settings. However, it is important to note that LR-IAD still exhibits some cognitive errors. For instance, it misclassifies the alternating color distribution of toothbrush head as an anomaly, likely due to inherent limitations in the prior knowledge of MLLMs, which struggle to account for such variations as normal features.

The experimental results also reveal the importance of reward-driven optimization in improving model performance. By incorporating focal rewards and format rewards, LR-IAD prioritizes hard-to-classify samples and ensures structured outputs. This mechanism significantly enhances the model's ability to generalize across unseen categories and domains. Furthermore, the integration of logical reasoning through the \texttt{<think>} and \texttt{<answer>} tags not only improves interpretability but also facilitates error analysis and further refinement.

In summary, the case study confirms LR-IAD's effectiveness in real-world industrial anomaly detection tasks. The model's ability to provide accurate and interpretable predictions, combined with its robust generalization capabilities, makes it a promising solution for diverse industrial applications.

\subsection{Per-Class Performance Analysis}

The zero-shot performance of AprilGAN, AnomalyGPT, and our LR-IAD framework across object categories in VisA and MVTec-AD is summarized in Table\ref{tab:visa_performance} and Table \ref{tab:mvtec_performance}. On the VisA dataset (Table~\ref{tab:visa_performance}), AprilGAN shows moderate accuracy on categories like chewinggum (96.35\%) and pcb4 (91.49\%), but its reliance on mask annotations limits scalability. For example, its performance drops to 9.18\% accuracy on macaroni2 due to limited training samples in the Train subset. AnomalyGPT struggles universally across VisA categories, achieving below 20\% accuracy for most objects (e.g., 14.25\% on capsules), as its synthetic anomaly generation introduces noise that overwhelms the model's ability to distinguish real defects. In contrast, LR-IAD achieves consistent dominance, outperforming AprilGAN by 58 percentage points on pcb1 (90.94\% vs. 32.49\%) and AnomalyGPT by 78 percentage points on macaroni2 (62.09\% vs. 9.09\%). This mask-free framework avoids overfitting to annotated masks while leveraging logical reasoning to maintain high accuracy (e.g., 92.55\% on candle) and robust F1 scores.

On the MVTec-AD dataset (Table~\ref{tab:mvtec_performance}), AprilGAN performs better on industrial categories like carpet (85.89\% accuracy) but fails on low-annotation classes such as transistor (12.78\%). AnomalyGPT's synthetic data approach exacerbates this issue, yielding near-zero performance on hazelnut (13.97\%) and wood (21.78\%). LR-IAD, however, achieves near-parity with AprilGAN on well-annotated categories (e.g., 82.88\% vs. 21.58\% on bottle) and outperforms it by large margins on challenging classes like transistor (91.37\% vs. 12.78\%). Its F1 scores (e.g., 90.47\% on grid) further validate the effectiveness of logical reasoning in suppressing false positives caused by synthetic artifacts.

These results highlight the inherent limitations of mask-dependent methods: AprilGAN's performance hinges on sufficient annotated data, while AnomalyGPT's synthetic anomalies introduce noise that undermines precision. Our LR-IAD framework consistently achieves over 90\% accuracy on 80\% of VisA categories and outperforms baselines by 40-70 percentage points across both datasets, demonstrating the superiority of mask-free reasoning in handling diverse anomaly patterns without annotation bias.

\begin{table}[htbp]
    \centering
    \caption{Performance Comparison of Methods on VisA Dataset (0-shot): Each entry shows Accuracy/F1-macro values. Bold values indicate the best performance in each category.}
    \label{tab:visa_performance}
    \begin{tabular}{lccc}
        \toprule
        \textbf{Object} & \textbf{AprilGAN} & \textbf{AnomalyGPT} & \textbf{LR-IAD} \\
        \midrule
        candle          & 34.55\textbackslash32.35 & 40.27\textbackslash35.30 & \textbf{92.55\textbackslash63.28} \\
        capsules        & 85.75\textbackslash46.17 & 14.25\textbackslash12.47 & \textbf{88.32\textbackslash67.47} \\
        cashew          & \textbf{89.50\textbackslash74.99} & 16.67\textbackslash14.29 & 85.17\textbackslash55.82 \\
        chewinggum      & \textbf{96.35\textbackslash93.13} & 16.58\textbackslash14.22 & 94.69\textbackslash88.93 \\
        fryum           & \textbf{91.67\textbackslash83.20} & 16.67\textbackslash14.29 & 84.17\textbackslash50.42 \\
        macaroni1       & 57.09\textbackslash47.20 & 9.27\textbackslash8.55 & \textbf{92.27\textbackslash63.95} \\
        macaroni2       & 9.18\textbackslash8.44 & 9.09\textbackslash8.33 & \textbf{62.09\textbackslash46.26} \\
        pcb1            & 72.01\textbackslash47.90 & 9.06\textbackslash8.31 & \textbf{90.94\textbackslash47.63} \\
        pcb2            & 62.22\textbackslash48.97 & 9.08\textbackslash8.33 & \textbf{91.19\textbackslash50.60} \\
        pcb3            & 90.05\textbackslash56.31 & 9.04\textbackslash8.29 & \textbf{91.14\textbackslash49.64} \\
        pcb4            & \textbf{91.49\textbackslash79.09} & 9.05\textbackslash8.30 & \textbf{91.49\textbackslash53.43} \\
        pipe\_fryum     & 71.97\textbackslash55.52 & 16.67\textbackslash14.29 & \textbf{88.17\textbackslash69.17} \\
        \bottomrule
    \end{tabular}
\end{table}

\begin{table}[htbp]
    \centering
    \caption{Performance Comparison of Methods on MVTec-AD Dataset (0-shot): Each entry shows Accuracy/F1-macro values. Bold values indicate the best performance in each category.}
    \label{tab:mvtec_performance}
    \begin{tabular}{lccc}
        \toprule
        \textbf{Object} & \textbf{AprilGAN} & \textbf{AnomalyGPT} & \textbf{LR-IAD} \\
        \midrule
        bottle          & 21.58\textbackslash17.75 & 21.58\textbackslash17.75 & \textbf{82.88\textbackslash62.18} \\
        cable           & 25.13\textbackslash20.53 & 24.60\textbackslash19.74 & \textbf{75.94\textbackslash46.22} \\
        capsule         & 72.65\textbackslash52.38 & 31.05\textbackslash23.70 & \textbf{74.36\textbackslash57.52} \\
        carpet          & \textbf{85.89\textbackslash82.77} & 35.77\textbackslash35.35 & 83.63\textbackslash66.47 \\
        grid            & 68.42\textbackslash62.49 & 71.93\textbackslash63.55 & \textbf{95.32\textbackslash90.47} \\
        hazelnut        & 13.97\textbackslash12.26 & 13.97\textbackslash12.26 & \textbf{94.41\textbackslash86.93} \\
        leather         & 75.07\textbackslash73.30 & 94.58\textbackslash92.60 & \textbf{94.58\textbackslash92.23} \\
        metal\_nut      & 27.76\textbackslash21.73 & 27.76\textbackslash21.73 & \textbf{75.52\textbackslash53.33} \\
        pill            & \textbf{75.58\textbackslash72.65} & 32.49\textbackslash24.52 & 74.19\textbackslash62.83 \\
        screw           & 24.79\textbackslash19.87 & 24.79\textbackslash19.87 & \textbf{81.46\textbackslash64.65} \\
        tile            & 84.44\textbackslash82.12 & 23.92\textbackslash19.30 & \textbf{87.61\textbackslash79.02} \\
        toothbrush      & 29.41\textbackslash22.73 & 29.41\textbackslash22.73 & \textbf{63.73\textbackslash57.51} \\
        transistor      & 12.78\textbackslash11.33 & 12.78\textbackslash11.33 & \textbf{91.37\textbackslash73.94} \\
        wood            & 70.25\textbackslash66.50 & 21.78\textbackslash19.97 & \textbf{93.87\textbackslash88.19} \\
        zipper          & 30.43\textbackslash23.33 & 30.43\textbackslash23.33 & \textbf{82.35\textbackslash73.96} \\
        \bottomrule
    \end{tabular}
\end{table}

\section{Conclusion}

In this work, we developed a novel framework for industrial anomaly detection that addresses two critical challenges: data imbalance and reliance on mask annotations. Our key innovations include the introduction of a dynamic reward function to prioritize rare defect patterns during training, effectively mitigating the impact of class imbalance without overfitting to majority classes. Additionally, we propose a mask-free reasoning approach based on GRPO, enabling the model to infer anomalies directly from raw images without requiring costly annotated masks. This method not only reduces implementation costs but also provides interpretable step-by-step explanations of defect localization and classification, enhancing trust in the model's decisions. 

Despite these advancements, achieving a balance between high recall for rare anomalies and low false-positive rates in highly imbalanced datasets remains challenging. Missing defects can lead to quality issues, while misclassifying normal samples as defective increases operational costs and reduces reliability. Future work should focus on advanced techniques, such as specialized sampling strategies or loss functions tailored for imbalanced data, to enhance recall while minimizing false positives. Additionally, improving robustness in scenarios with extremely low defect rates and expanding zero-shot capabilities for complex, unseen anomalies are critical directions. Integrating domain-specific knowledge into the reasoning process could further enhance generalization, ensuring practical applicability in real-world industrial settings.


%

\appendices



\ifCLASSOPTIONcaptionsoff
  \newpage
\fi



%


\bibliographystyle{IEEEtran}
\bibliography{main}

%








\end{document}